\documentclass[review]{elsarticle}

\usepackage{lineno,hyperref}
\usepackage{graphicx}
\graphicspath{{Figure/}}        
\usepackage{times,amsmath,epsfig}
\usepackage{subfigure}
\usepackage{multirow}
\usepackage{bm}
\usepackage{amsmath}
\usepackage{booktabs}
\usepackage{bigstrut}
\usepackage{amssymb}
\usepackage{setspace}
\usepackage{lineno}
\usepackage{color}
\usepackage{ulem}
\usepackage{tabularray}
\usepackage{parskip}
\usepackage{caption}
\usepackage{multicol}
\usepackage{xcolor}
\usepackage{subcaption}
\usepackage{subcaption}

\biboptions{authoryear}

\begin{document}
\begin{frontmatter}

\title{Enabling CMF Estimation in Data-Constrained Scenarios: A Semantic-Encoding Knowledge Mining Model}
\author[mymainaddress]{Yanlin Qi}
\author[addressb]{Jia Li}
\author[mymainaddress]{Michael Zhang\corref{mycorrespondingauthor}}
\cortext[mycorrespondingauthor]{Corresponding author}
\ead{hmzhang@ucdavis.edu}

\address[mymainaddress]{Institute of Transportation Studies, University of California, Davis, CA 95616, USA}
\address[addressb]{Department of Civil and Environmental Engineering, Washington State University, WA 99164, USA}

\begin{abstract}

Precise estimation of Crash Modification Factors (CMFs) is central to evaluating the effectiveness of various road safety treatments and prioritizing infrastructure investment accordingly. 
While customized study for each countermeasure scenario is desired, the conventional CMF estimation approaches rely heavily on the availability of crash data at given sites. This not only makes the estimation costly, but the results are also less transferable, since the intrinsic similarities between different safety countermeasure scenarios are not fully explored. Aiming to fill this gap, this study introduces a novel knowledge-mining framework for CMF prediction. This framework delves into the connections of existing countermeasures and reduces the reliance of CMF estimation on crash data availability and manual data collection. Specifically, it draws inspiration from human comprehension processes and introduces advanced Natural Language Processing (NLP) techniques to extract intricate variations and patterns from existing CMF knowledge. It effectively encodes unstructured countermeasure scenarios into machine-readable representations and models the complex relationships between scenarios and CMF values. This new data-driven framework provides a cost-effective and adaptable solution that complements the case-specific approaches for CMF estimation, which is particularly beneficial when availability of crash data or time imposes constraints. Experimental validation using real-world CMF Clearinghouse data demonstrates the effectiveness of this new approach, which shows significant accuracy improvements compared to baseline methods. This approach provides insights into new possibilities of harnessing accumulated transportation knowledge in various applications.

\end{abstract}

\end{frontmatter}

\section{Introduction}

Road safety stands as a critical concern affecting the well-being and safety of all road users. In the pursuit of enhanced sustainability and reduced crashes, transportation agencies allocate substantial resources towards implementing a wide array of countermeasures within the road infrastructure. Like many other fields, road safety professionals rely on quantitative assessments, including crash modification factors (CMFs), as a crucial metric to evaluate the effectiveness of safety countermeasures in reducing crashes \citep{national2010highway}. This information is crucial for making informed decisions regarding the implementation of these safety measures. \citep{yannis2016good}. 

In this context, data-driven safety analysis (DDSA) has been an established field in road safety research. Road safety professionals have extensively used statistical analyses of crash data to assess various countermeasure scenarios \citep{wunderlich2020data}.  These efforts have yielded numerous case-specific CMFs that provide quantitative insights into expected crash rates after implementing countermeasures. The hope is that through intensive efforts, all known countermeasures can be covered, thereby more efficient safety countermeasures can be put into place to reduce future road crashes.

Unfortunately, existing CMF coverage does not adequately address the wide range of countermeasure scenarios for several reasons. Firstly, the field continually witnesses the introduction of new countermeasures, resulting in a constantly evolving landscape.  Additionally, the effectiveness of any given countermeasure can vary depending on contextual factors such as geography, site conditions, crash types, and severity levels \citep{souleyrette2020crash}. These factors contribute to a multitude of potential countermeasure scenarios. For example, while a CMF may exist for roundabout conversions based on previous research, it may be specific to rural, stop-controlled intersections. This leaves clear gaps in the coverage of the CMF, such as the absence of a CMF applicable to the conversion of signalized rural intersections \citep{gross2015crash}. This diversity in countermeasure scenarios necessitates the ongoing development of CMFs for potential unexplored applications. While customizing research to create precise CMFs for each unique road safety treatment and condition is ideal and dependable,  the challenge of broadening the available CMFs to cover a wide range of potential safety treatment scenarios remains substantial.

A significant challenge arises from the high cost of transferring the prediction models across diverse countermeasure scenarios. Various statistical and data mining techniques, including before-after approaches \citep{gross2010guide, abdel2016validation}, empirical Bayesian approaches \citep{elvik2017empirical},  multivariate regression models \citep{wu2016investigating, sacchi2014collision, park2015developing}, and more advanced machine learning approaches \citep{park2015assessing, hauer2004safety}, have been employed in the field. However, these data-driven analyses are primarily tailored to specific countermeasure scenarios and are often associated with high costs and time requirements \citep{la2016development, wu2017developing}. This stems from the fact that, regardless of the statistical models used, data collection involves two main types: before-after and cross-sectional data on crash inventories \citep{gross2010guide}. Before-after studies require the collection of multiyear crash data from a single group or location both before and after a particular intervention \citep{park2015developing}.  In contrast, cross-sectional studies involve the coordination of crash data collection from multiple groups or locations at a single point in time \citep{wu2017developing}. Regardless of the approach, careful planning and execution are essential to ensure the relevance and reliability of the collected data for statistical analysis of treatment effectiveness \citep{wood2015comparison}. Even with the introduction of recent machine learning models for CMF development,  the need for repeated data collection and model development persists when addressing different countermeasure scenarios. Consequently, achieving an estimate of safety effectiveness through the implementation of treatment and the collection of crash data in the short term may be challenging.


Statistical analysis of crash data may even not be feasible in data-limited scenarios. CMFs are typically expressed as a ratio of the expected number of crashes with countermeasure compared to without. Consequently, researchers often depend on statistical analysis of crash inventories to develop CMFs, making crash databases pivotal \citep{imprialou2019crash}. This reliance on crash data poses a significant obstacle since these methods demand the availability of crash inventory data. This requirement restricts the evaluation of effectiveness before implementing a treatment or accumulating a sufficient number of crashes for analysis \citep{al2023review}. One shared limitation these methods have is that they adhere to the conventional data-mining strategy that solely focuses on mining numerical data and crash observations.

In practice, when existing crash data are insufficient and time and resources are limited for collecting more data, safety professionals often leverage their judgment to retrieve valuable patterns and contexts from previous CMF studies. For instance, when assessing an uninvestigated countermeasure scenario, safety practitioners may infer its safety effectiveness by comparing it to similar investigated cases. This process involves comparing semantic countermeasure descriptions and contextual factors to estimate the CMF value for the target scenario based on expert judgment. Although safety experts frequently use prior knowledge of CMFs in this intuitive manner, this process is neither systematic nor reproducible and can be challenging when dealing with complex situations. An alternative method that has been proposed is meta-analysis \citep{bahar2010methodology, gross2011investigation}, which uses statistical techniques to synthesize and analyze results from multiple individual studies that estimate CMF for specific safety countermeasures. However, this method has limitations, such as human's subjectivity, not fully harnessing the entire CMF knowledge base, and lacking transferability. 

To address these challenges, we introduce a knowledge-mining CMF prediction framework that alleviates the reliance on crash data and manual data collection, facilitating seamless applicability across diverse countermeasure scenarios.  
Our approach draws inspiration from human comprehension processes and leverages advanced natural language processing (NLP) techniques to extract intricate variations and patterns from existing CMF knowledge. It discerns key elements distinguishing different countermeasures, contextual factors, and their corresponding CMF values. The framework effectively encodes unstructured countermeasure scenarios into machine-readable representations and models the complex relationship between these scenarios and their associated CMF values. 
This approach proves especially valuable in scenarios with limited crash data or stringent time constraints, complementing existing case-specific CMF prediction methods. Furthermore, it has the potential to unlock previously untapped opportunities for utilizing the wealth of accumulated transportation knowledge in various applications.
In summary, the main contributions of this study are as follows.
\begin{itemize}
    \item \textbf{Cost-effective CMF estimation.} We introduce a novel approach that formulates CMF estimation as a knowledge-mining problem. This method maximizes the use of previously accumulated CMF data, resulting in a cost-effective estimation process seamlessly adaptable to various countermeasure scenarios. It complements traditional methods and enables rapid and efficient CMF predictions, particularly when dealing with limited crash data or time constraints.

    \item \textbf{Data-driven framework.} We present a data-driven framework for mining unstructured data. This framework leverages cutting-edge NLP techniques to effectively capture and model the complex relationships between countermeasure scenarios and CMF values. It adeptly captures the semantic contexts associated with diverse countermeasure scenarios while handling high-cardinality categories, addressing missing data, and managing noisy information. 
    
    \item \textbf{Experimental validation.} We apply our approach to real-world CMF Clearinghouse data, conducting extensive experiments and analysis. Comparing it with two baseline methods, our results demonstrate significant accuracy improvement. The framework proves its ability to predict CMFs with reasonable accuracy.

\end{itemize}

\section{Related work}
\subsection{The usage of CMF in transportation}

In road safety, CMF is a widely-used metric for evaluating the effectiveness of safety treatments on crash reduction. CMF values greater than 1.0 indicate an expected increase in crashes following the implementation of a specific countermeasure, whereas values less than 1.0 suggest a reduction in crashes \citep{national2010highway}. For a specific countermeasure, its safety effectiveness may be measured by a CMF applicable for all crash types and locations (e.g., all crashes for all area types) or multiple CMFs for each of which only measures for a specific contextual condition (e.g., angle crashes at rural signalized intersections). The applicability of a CMF depends upon the underlying study from which the CMF was estimated. The population of CMF studies are constantly expanding and some new countermeasures may not have been investigated yet. To facilitate access to CMFs, the Federal Highway Administration (FHWA) maintains the CMF Clearinghouse \citep{jones2010cmf}, serving as a repository for this valuable prior knowledge of CMF. Researchers from the U.S., Canada, and other parts of the world contribute to this repository, resulting in multiple CMFs associated with a single countermeasure.

CMFs are used extensively among various professionals involved in road safety. They serve as crucial metrics for evaluating the effectiveness of safety countermeasures, and are instrumental in cost-benefit analyses of safety interventions \citep{harkey2008accident}.  Specifically, they aid in assessing the safety impacts of different countermeasures and enable comparisons among various options and locations \citep{national2010highway}. Further, CMFs can be used to categorize cost-effective strategies and locations based on their effects on reducing crashes \citep{gross2010guide}.

\subsection{CMF prediction}
The literature has employed several methodologies to estimate CMFs in a case-specific way. The naive before-after method, a straightforward approach, involves comparing crash rates before and after implementing a treatment \citep{graham1982effectiveness, pitale2009benefit}. Although this approach is easy to use,  it has limitations in accounting for factors that might influence crash rates. To enhance CMF estimation,  the before-after approach with comparison group can be used by using reference sites similar to the treatment site \citep{retting2002changes, de2005road, noyce2004safety}. However, for this method to be effective, the treatment must be randomly implemented, and reference sites must closely resemble the treatment site, including crash rates in the before period.  These conditions are often not met, particularly since countermeasures tend to be applied at high-risk sites, introducing self-selection bias. Additionally, selecting suitable reference sites can be challenging, and in some cases, the treatment may also impact crash rates at nearby reference sites.

The empirical Bayesian before–after approach is currently the most widely used methodology, designed to address the regression to the mean effect \citep{harkey2008accident, park2012safety, persaud2013evaluation}. This method uses reference sites to estimate the expected number of crashes that would have occurred without treatment.  Negative binomial models are typically employed to estimate safety performance functions (SPF) for crash prediction, which provide expected crash rates and variances \citep{yannis2017road}. 

CMFs can also be developed from multivariate regression models of crash rates, using a set of explanatory variables such as traffic volume, segment length, and geometric design \citep{cafiso2010development, labi2011efficacies, turner2012next, persaud2013evaluation}. These models commonly use Poisson \citep{wichert2007accident} or negative binomial \citep{wu2008accident, persaud2013evaluation, al2021development}  regression. While multivariate regression models can be useful when only cross-sectional data are available, it does not take into account of the non-random nature of treatment implementation.  As a result, more advanced modeling techniques such as instrumental variables are required to obtain unbiased estimates of treatment effect \citep{karathodorou2016development}. 

More recently, machine learning (ML) approaches have also been employed for CMF estimations \citep{al2023review}, which have shown promising performance in providing higher predictive accuracy of CMFs. Some explainable ML models such as MARS, SHAP have also been used to derive explainable CMFs \citep{wen2022interpretability}. ML methods offer remarkable  flexibility, requiring minimal or no prior assumptions regarding crash data, and they can handle noises and outliers \citep{tang2019crash}. However, current ML-based studies mainly focus on developing case-specific machine learning models, which are not transferable between different countermeasure scenarios. Additionally, these data-driven machine learning models demand extensive crash data for meaningful training. For instance, Wen et al. have applied the light gradient boosting machine (LightGBM) and the shapley additive explanation (SHAP) to derive CMFs for run-off-road (ROR) crashes based on nearly tens of thousands of crash records in Washington State. 

\subsection{Knowledge mining}

In the field of data analysis, knowledge mining has emerged as a powerful methodology for extracting valuable insights from vast datasets and textual information \citep{nasukawa2001text}.  It encompasses techniques like data mining, text mining, and information retrieval to transform raw data into actionable knowledge, facilitating informed decision-making.

Within the transportation domain, the fusion of these methodologies has led to innovative research \citep{erfani2021empirical, lock2020social, gu2016twitter}.  For instance, a study by \cite{rath2022worldwide} introduces a novel approach that leverages advanced NLP and Wikipedia data to predict city transport typologies. This method effectively addresses data scarcity challenges and leverages text-based information sources to benefit urban planning and transportation research. Similarly, \cite{gu2016twitter} present a method for mining tweet texts to detect real-time traffic incidents, thus providing a cost-effective complement to traditional data sources.

In the context of road safety, researchers have applied these methodologies to unveil hidden patterns \citep{zhong2020hazard, srinivas2023passenger, francis2023smartxt}. \cite{ahadh2021text} introduced a semi-supervised method for crash reports to reduce the dependence on extensive labeled data. Their approach, which incorporates domain-specific keywords and topics, achieves an average accuracy of 80\%  without requiring extensive labeled data.  In a similar vein, the investigation by \cite{aryffin2021public} used Twitter data to gauge public perceptions of road safety. Employing supervised learning,  \cite{janstrup2023predicting} leverage text mining and neural network regression to anticipate injury severity distributions in cyclist crashes. The extraction of crucial safety insights from textual data and crash records highlights the potential of these approaches.

Literature mining, a subset of knowledge mining, specializes in extracting pertinent information and insights from scholarly literature and research reports \citep{jensen2006literature, papanikolaou2014biotextquest}. By utilizing natural language processing and text mining, literature mining extracts valuable information and insights embedded within textual sources.  The work of \cite{toor2022didace} serves as an illustrative example of literature mining's potential in unveiling connections between diseases and diets, a task achievable through in-depth analysis of biomedical research papers.

While some research has delved into text understanding-based techniques for crash-related predictions and other transportation applications, the exploration of knowledge mining remains in largely uncharted territory within the transportation domain. In our work, we synthesize these principles and leverage the critical information extracted from CMF literature to predict countermeasure effectiveness.
\section{Problem formulation}
\subsection{Overview}
In this section, we formally define the knowledge-driven CMF prediction problem. Our goal is to develop a knowledge-driven CMF prediction framework capable of efficiently mining and encoding the complex variations and patterns within existing CMF studies. This framework aims to predict CMFs for new, previously unseen countermeasure scenarios by leveraging the semantic context and pattern variations learnt among countermeasure descriptions and contextual factors. 

\subsection{Setup}
Our approach considers each unique CMF study as an individual countermeasure scenario, extracting key elements from the CMF study records archived in the FHWA CMF Clearinghouse \footnote{\url{https://www.cmfclearinghouse.org/results.php}}. These elements encompass crucial information related to countermeasure descriptions (e.g., converting angle parking to parallel parking), crash details (e.g., parking-related crashes), geographical attributes (e.g., urban areas), infrastructure configurations (e.g., 1 or 2-lane, local roads), and investigated CMF values (e.g., 0.37). 

To elaborate further, we assemble a set $S$ of CMF studies from the FHWA CMF Clearinghouse, where each research record $s \in S$ corresponds to a unique countermeasure scenario. Each countermeasure scenario $s_i$ involves a specific countermeasure ($cm$) and multiple circumstantial factors ($\mathbf{u} = [u_{1}, u_{2}, ..., u_{k}]^T\in \mathbb{R}^{k\times1}$) examined within a CMF study. We distinguish between scenarios with known CMF values and those requiring CMF estimation as known and unknown countermeasure scenarios, respectively.
\subsection{Objective}
The first-stage objective of our work is to efficiently convert each unstructured countermeasure scenario ($s$) into a machine-readable vector-based representation ($\mathbf{v}=feat\_encoder(s)$) by developing a feature encoder. Importantly, the similarity between the feature representations ($vec\_simi(\mathbf{v}_i, \mathbf{v}_j)$) of any countermeasure scenario pair ($s_i$ and $s_j$) should effectively reflect their similarity in safety performance and effectiveness ($safety\_simi(s_i, s_j)$), as given Equation \ref{eq.encoder_obj}:

\begin{equation}
        {\min} \Sigma_1^{N_p}[vec\_simi(\mathbf{v}_i, \mathbf{v}_j)
            -safety\_simi(s_i, s_j)]^2
\label{eq.encoder_obj}
\end{equation}
where ${N_p}$ is the number of countermeasure scenario pairs.

The second-stage objective is to develop a predictive model ($cmf\_regr$) that can establish the complex connection between the embedding representations and the corresponding CMF values. This model should estimate CMF values for unknown countermeasure scenarios by drawing upon knowledge from similar known scenarios. To achieve this, we train a CMF regressor using supervised learning for regression, optimizing the mean-squared-error (MSE) loss function as defined in Equation \ref{eq.loss_func}:

\begin{equation}
    \min \space \frac{1}{N}\Sigma [{cmf_i}-cmf\_regr(\mathbf{v}_i)]^2
    \label{eq.loss_func}
\end{equation}
where $N$ is the sample size of investigated known countermeasure scenarios.
\section{Method}
\subsection{Framework architecture}
\begin{figure*}[!ht]
  \centering
  \includegraphics[width=1.0\textwidth]{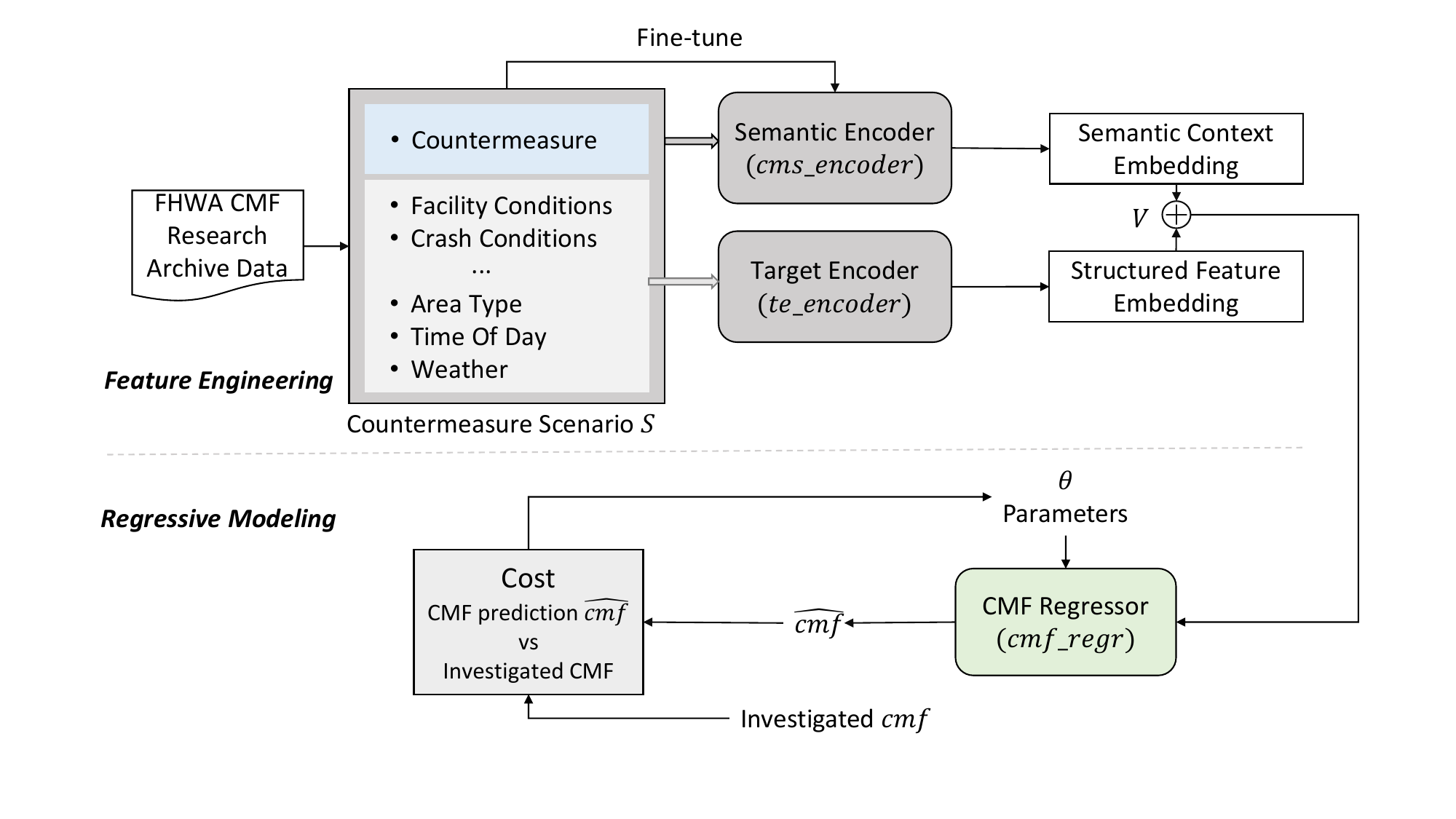}
  \caption{The proposed countermeasure scenario encoding-based CMF prediction framework} 
  \label{fig:cmfframework}
\end{figure*}

Figure \ref{fig:cmfframework} illustrates the architecture of our proposed framework. In the first stage, we introduce a countermeasure scenario semantic encoder $cms\_encoder$ to capture unstructured semantic contexts and a target encoder $te\_encoder$ to capture high-dimensional contextual elements. Both encoders play crucial roles in transforming machine-unreadable countermeasure scenarios into vector representations.  To enhance the efficiency of the semantic encoder, we employ an adaptation strategy to fine-tune the $cms\_encoder$ for domain adaptation. In the second stage, we train the MLP regression model, which is responsible for learning and establishing the complex relationship between the vector representations and their corresponding CMF values.

\subsection{Data description}

We adopt the FHWA CMF Clearinghouse \citep{jones2010cmf} as our primary CMF knowledge database, which is the online repository that dynamically aggregates CMFs from scholarly efforts.  At the completion of this work, it contains a total of 9,307 CMF records, along with associated countermeasures and site conditions. This repository includes various data fields that describe CMF applicability, such as countermeasure names, categories, infrastructure and area types, crash details,  method specifics, study details (e.g., title and publish year), and statistical properties (e.g., quality star rating).  However, due to inherent data characteristics and limitations, it predominantly comprises unstructured data with significant missing values. For instance, countermeasures are often described in narrative form, conveying critical contextual information, as exemplified in Table~\ref{tab.struvsunstru}.

\begin{table}
\centering
\caption{Statistics of the structured and unstructured countermeasures categorized by the roadway segment type and intersection type}
\label{tab.struvsunstru}
\noindent\begin{minipage}{\textwidth}
\resizebox{\textwidth}{!}{%
\begin{tblr}{
width = \textwidth,
  hline{1,4} = {-}{0.08em},
  hline{2} = {-}{},
}
Facility Type & Countermeasure Example                                 & Count & Percentage \\
Roadway       & Reduce lane width from 12 ft to less than 12 ft        & 5871  & 63.07\%    \\
Intersection  & Painted channelization of left-turn lane on major road & 3436  & 36.93\%    
\end{tblr}}
\end{minipage}
\end{table}

\begin{figure*}[!ht]
    \centering
    \subfigure[Roadway]{
        \includegraphics[width=0.48\linewidth]{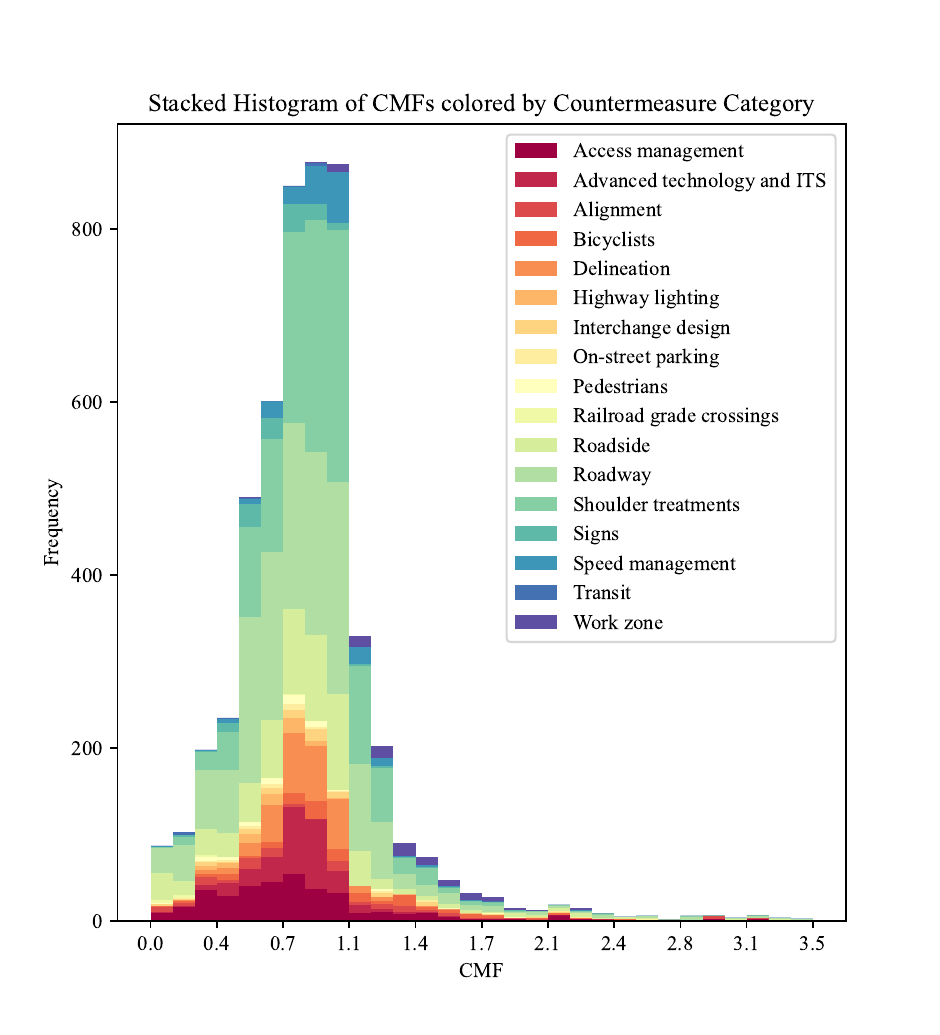}
    }
    \subfigure[Intersection]{
	\includegraphics[width=0.48\linewidth]{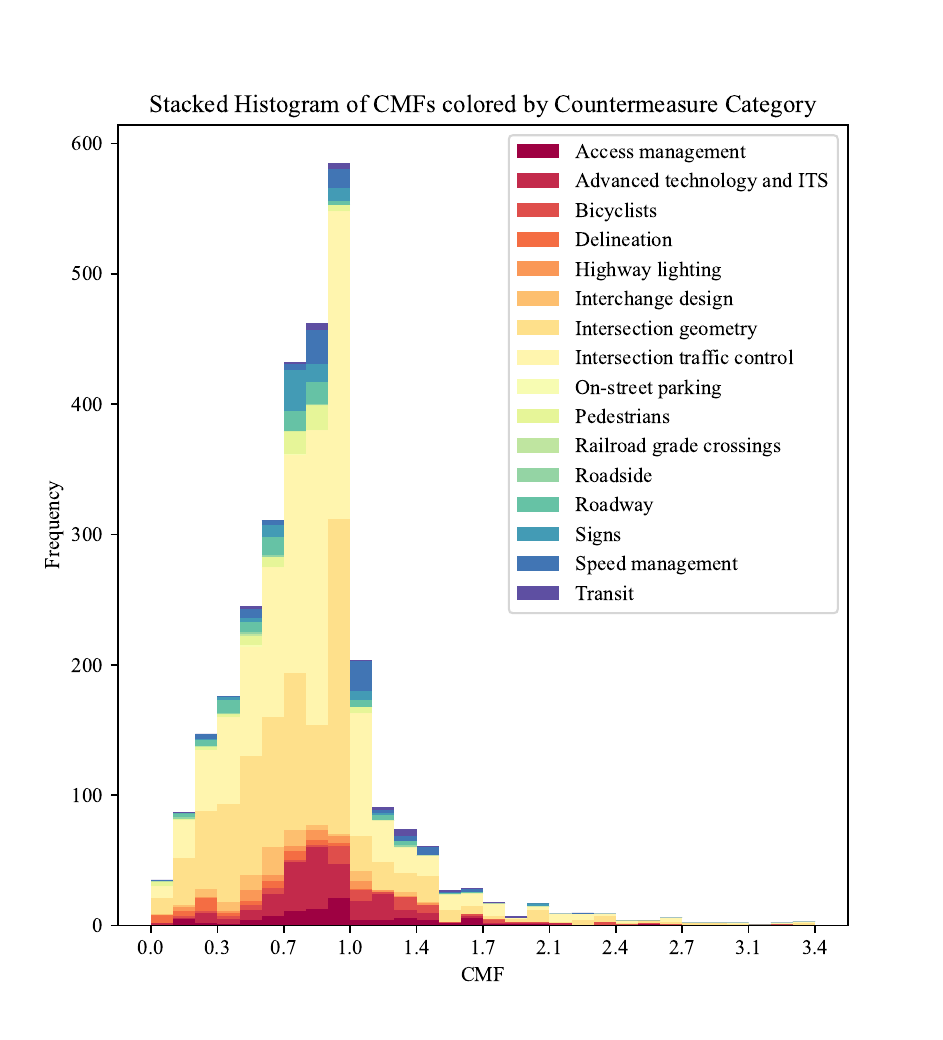}
    }
    \caption{The CMF frequency distribution across various countermeasure categories: (a) roadway segment  facility type and (b) intersection facility type in the FHWA CMF clearinghouse }
    \label{fig.catdis}
\end{figure*}

Scholarly efforts have contributed a diverse collection of investigated countermeasures spanning various categories, which provides the data foundation for our knowledge-mining approach. Figure \ref{fig.catdis} illustrates the frequency distribution of CMF values among these categories, highlighting the richness of the data set.  Notably, the most frequently studied categories for intersection countermeasures include \textit{Intersection geometry} and \textit{intersection traffic control}, while for roadway segments, \textit{Roadway} and \textit{Shoulder width} dominate.
The distribution of CMF values reveals that the majority of countermeasures have values close to $1.0$, indicating a relatively minor safety impact. This aligns with the common goal of many safety countermeasures to maintain or slightly improve safety conditions. Additionally, most CMF values fall within the range of $[0, 2.0]$, suggesting that the majority of countermeasures operate within this span of safety effectiveness.  To prevent overfitting to extreme and less common cases, countermeasures with CMF values exceeding 2.0 are excluded as statistical outliers.

The repository categorizes countermeasures into two main types: roadway segments and intersections, aligned with site condition geometries. To address the issue of missing data fields (e.g., missing intersection type for certain roadway segment countermeasures), we adopt a consistent framework and strategy to separately train and evaluate our model for roadway-related and intersection-related countermeasure scenarios.  This strategy reduces unnecessary input demands and minimizes deviations.

\subsubsection{Key features defining countermeasure scenarios}

Known influential factors that affects the safety performance of countermeasure safety effectiveness \citep{gross2010guide} are identified from the original CMF clearinghouse data fields.These factors encompass information related to countermeasures, crashes, geographical conditions, data time spans, and site conditions for each countermeasure scenario. We have organized the chosen data fields into six categories, as outlined in Table~\ref{tab:sta}.

The first four categories cover different aspects of countermeasures, crashes, area types, and data time spans. These categories encompass general influential factors contributing to CMF variations across different scenarios. The last two categories focus on subattributes specific to facility types: intersections and roadway segments. 
We treat these facility-related variables separately when developing models for each facility type. This approach is rooted in the observation that intersections and roadway segments often exhibit distinct characteristics associated with countermeasures.

With these chosen features in hand, our next step is to create a streamlined and effective predictive modeling approach that captures the variation patterns among different countermeasure scenarios. 

\begin{table}
\centering
\caption{Overview of chosen data fields for CMF prediction}
\label{tab:sta}
\begin{tblr}{
  width = \linewidth,
  colspec = {Q[190]Q[312]Q[167]Q[148]Q[123]},
  cell{2}{1} = {r=3}{},
  cell{5}{1} = {r=3}{},
  cell{8}{1} = {r=3}{},
  cell{11}{1} = {r=2}{},
  cell{13}{1} = {r=4}{},
  cell{17}{1} = {r=4}{},
  hline{1,21} = {-}{0.08em},
  hline{2} = {-}{},
}
Feature Category & Variable                     & {Missing Rate \\ for Intersection} & {Missing Rate\\ for Roadway} & {Number of\\ Categories} \\
Countermeasure   & Countermeasure name          & 0.0\%                              & 0.0\%                        & -                        \\
                 & Countermeasure category      & 0.0\%                              & 0.0\%                        & 19                       \\
                 & Countermeasure subcategory   & 0.0\%                              & 0.0\%                        & 40                       \\
Crash            & Crash type                   & 9.1\%                              & 15.4\%                       & 100                      \\
                 & Crash time-of-day            & 17.7\%                             & 21.3\%                       & 5                        \\
                 & Crash severity               & 0.4\%                              & 16.5\%                       & 19                       \\
Local area       & Area type                    & 14.7\%                             & 23.2\%                       & 7                        \\
                 & Country                      & 33.1\%                             & 21.2\%                       & 27                       \\
                 & State/City                   & 20.5\%                             & 18.5\%                       & 129                      \\
Time span        & Start year                   & 38.4\%                             & 24.6\%                       & -                        \\
                 & End year                     & 38.4\%                             & 24.6\%                       & -                        \\
Intersection     & Intersection type            & 6.8\%                              & -                            & 8                        \\
                 & Intersection geometry        & 12.8 \%                            & -                            & 9                        \\
                 & Traffic control type         & 7.6\%                              & -                            & 8                        \\
                 & Intersection prior condition & 29.9\%                             & -                            & -                        \\
Roadway          & Roadway type                 & -                                  & 51.3\%                       & 13                       \\
                 & Road division type           & -                                  & 16.5\%                       & 7                        \\
                 & Number of lanes              & -                                  & 41.4\%                       & 47                       \\
                 & Roadway prior condition      & -                                  & 39.0\%                       & -                        
\end{tblr}
\end{table}

\subsection{Feature encoding for countermeasure scenarios}

\subsubsection{Contextual semantic encoding}

As extracted features in their original form contain unstructured or semi-structured information, it is challenging to extract and learn meaningful patterns from them. Therefore, we concatenate the selected data fields for each countermeasure scenario to create a single text sequence, namely a \textit{ pseudo sentence}. We use comma separators to distinguish between different data fields within the \textit{pseudo sentence}. The key is to create \textit{pseudo sentences} that effectively convey the essential details of each countermeasure scenario while maintaining a consistent format for easy processing and analysis. These pseudo sequences will serve as the input data for further semantic encoding purpose.

To achieve semantic encoding of diverse countermeasure scenarios,  we selected the language model Sentence-BERT \citep{devlin2018bert, reimers2019sentence} as backbone, which produces basic sentence embeddings capturing general context and semantics. As Sentence-BERT is pretrained with general-purpose corpora, the raw backbone would not work well directly on countermeasure scenarios because of the domain shift. We thus fine-tune the Sentence-BERT using \textit{pseudo sentences} formed by the text sequence of the selected features for each countermeasure scenario.

\begin{figure}[htb]
        \centering
        \includegraphics[width=0.9\textwidth]{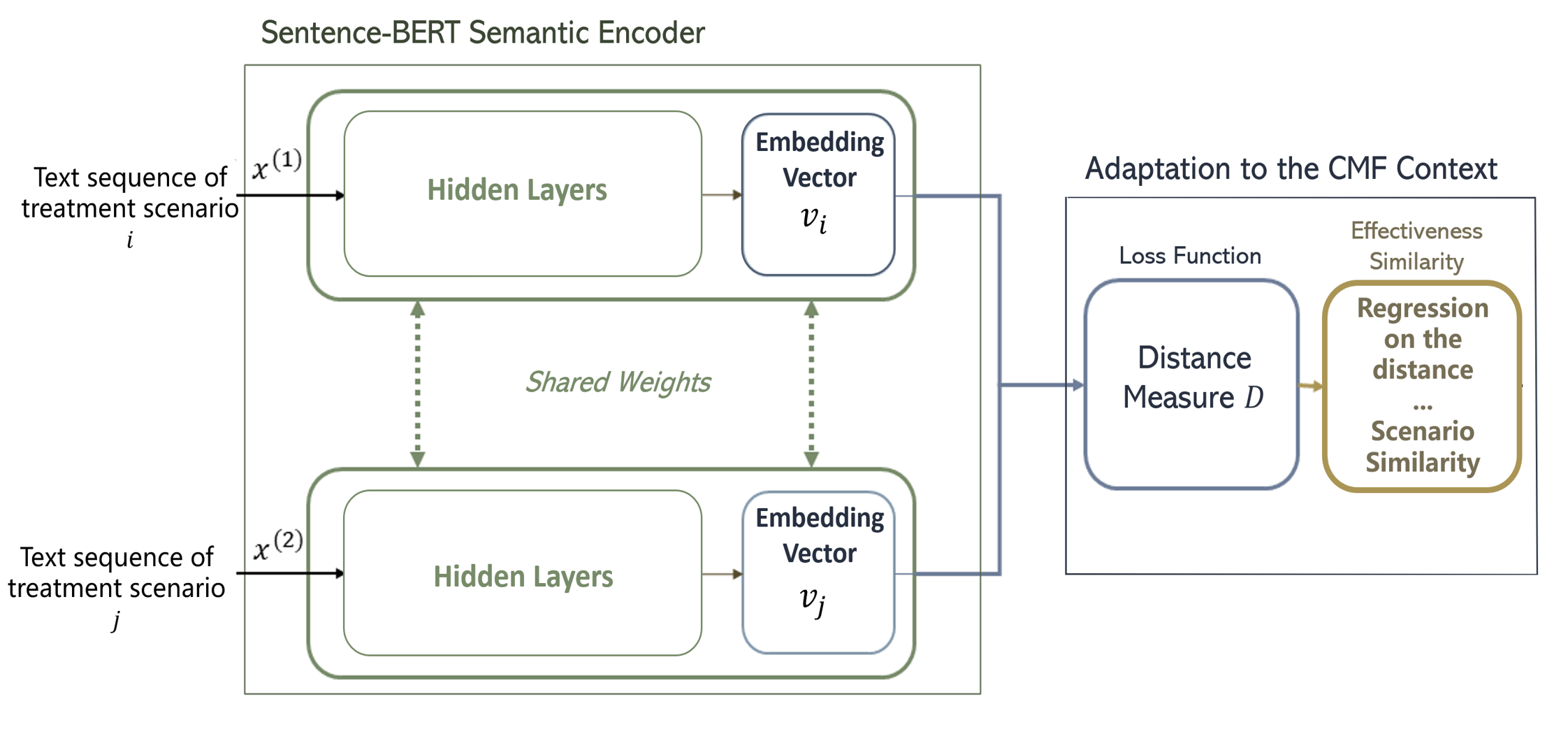}
        \caption{The fine-tuning for countermeasure scenario encoder from the pretrained Sentence-BERT language model}
        \label{fig:semantic}
\end{figure}

For domain adaptation, Sentence-BERT functions as a Siamese network (depicted in Figure \ref{fig:semantic}). It takes two \textit{pseudo sentences} as input and a domain-specific similarity score to adapt the corresponding vector representations in a high-dimensional space. These embeddings are adjusted to be meaningful by comparing the model-predicted vector-based similarity score with the degree of safety performance similarity between the two countermeasure scenarios. The latter requires a substantial dataset of countermeasure scenario pairs, each assigned a similarity score based on their safety performance similarity, a task which can be labor-intensive and challenging given the complex interplay of contextual factors and semantic nuances. 

To address this challenge, our first step involves creating a Safety Performance Similarity Function (SPSF) to replace manual labeling. In constructing the SPSF, we consider two critical factors:
\begin{itemize}
    \item \textbf{Nature of countermeasure scenarios.} Scenarios with CMF values above or below 1 are grouped as positive or negative, as they share similar effectiveness natures. Conversely, scenario pairs with one value above 1 and the other below 1 should be considered less alike due to differing effectiveness natures.
    
    \item \textbf{Degree of CMF value closeness.} This factor quantifies the numerical differences between the two CMF values.
\end{itemize}

\begin{figure}[!ht]
    \centering
    \includegraphics[width=0.4\textwidth]{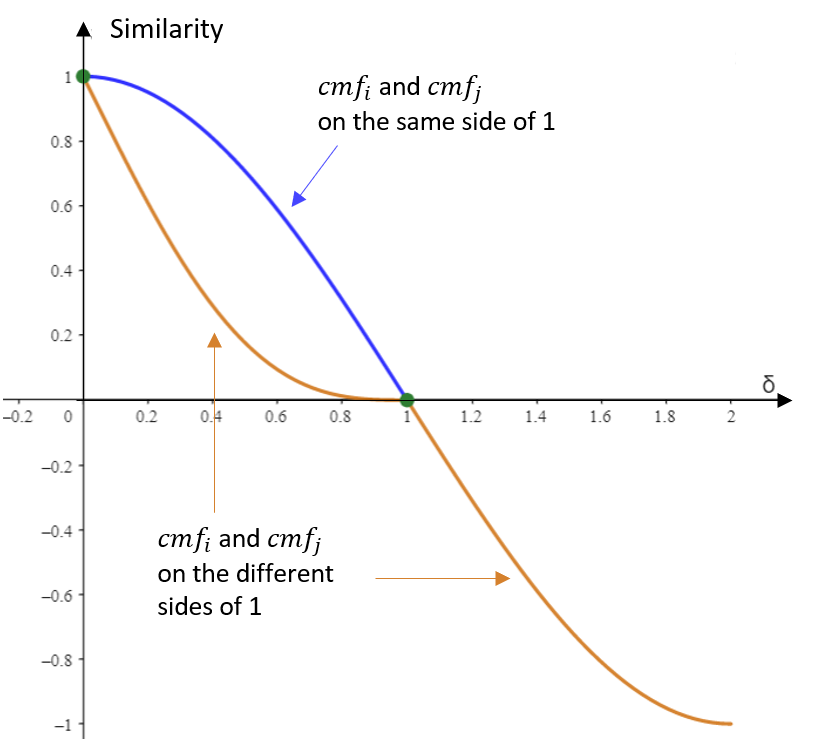}
    \caption{Illustration of the similarity score of a known countermeasure scenario pair $(s_i, s_j)$}
    \label{fig:simiscore}
\end{figure}

To comprehensively incorporate both aspects, we define a similarity score for two countermeasure scenarios using Equation \ref{eq:spsf}. It is important to note that the SPSF is designed specifically for domain adaptation and does not assess the exact safety effectiveness similarity between countermeasure scenarios.

\begin{equation}
\resizebox{0.9\linewidth}{!}{$\operatorname{Similarity}(s_i, s_j)=\left\{\begin{matrix}
\cos(\frac{\pi }{2} \delta)(\delta -1)^2, & if \quad \operatorname{Sgn}(cmf_i-1) \operatorname{Sgn}(cmf_j-1)\delta \in (-1, 0)\\ 
 \cos(\frac{\pi }{2} \delta),&otherwise 
\end{matrix}\right.$}
\label{eq:spsf}
\end{equation}

where $s_i, s_j$ represent the \textit{pseudo sentences} of any given countermeasure scenario $i$ and $j$ for comparison, and $cmf_i, cmf_j$ correspond to their respective investigated CMF values. We calculate the absolute difference of CMFs as $\delta=|cmf_i-cmf_j|$. This similarity score is bounded within the range of $[-1, 1]$ that aligns with the cosine similarity requirements. This concept is visualized in Figure \ref{fig:simiscore}. 

To measure the similarity between the embedding vectors of model output, we use the cosine similarity method to quantify the similarity, quantified by Equation \ref{eq.cosine}.

\begin{equation}\label{eq.cosine}
\begin{aligned}
     Cos\theta &=\frac{cms\_encoder(pseudo\_sent(s_i))\times cms\_encoder(pseudo\_sent(s_j))}{Norm(cms\_encoder(pseudo\_sent(s_i)))\times Norm( cms\_encoder(pseudo\_sent(s_j)))} \\
         &= \widetilde{cms\_encoder(pseudo\_sent(s_i))}^T \times \widetilde{cms\_encoder(pseudo\_sent(s_j))}
\end{aligned}
\end{equation}

Here, $cms\_encoder(pseudo\_sent(s)) \in \mathbb{R}^{m \times 1}$ represents the semantic encoding of the countermeasure scenario $s$. The cosine similarity computes the inner product of normalized vectors, ensuring that both vectors have a unit $l_2$ norm. The normalized vectors are represented as $\widetilde{cms\_encoder(pseudo\_sent}(s))$, and $T$ denotes the transpose operation. In essence, by calculating the matrix multiplication of the two normalized vectors $\widetilde{cms\_encoder(pseudo\_sent}(s_i))^T$ and $\widetilde{cms\_encoder(pseudo\_sent(s_j))}$, we derive the cosine similarity between the vectorized representations of the countermeasure scenarios  \\
($cms\_encoder(pseudo\_sent(s_i))$ and $cms\_encoder(pseudo\_sent(s_j))$). This similarity measure is then compared with the gold similarity score, and the fine-tuning process enhances the recognition of similarity between different countermeasure scenarios.


\subsubsection{Circumstantial factor encoding}

In addition to encoding the semantic context of the countermeasure scenarios, we enrich the input by including the encoding of circumstantial factors. These factors, extracted from semi-structured texts in the CMF Clearinghouse, offer valuable insights into the road environments where safety countermeasures are deployed. This addition aims to enhance the CMF prediction model's ability to capture subtle variations in safety impacts across different contexts.

However, integrating these structured features into the model presents challenges. High-cardinality attributes, such as specific countermeasure categories or crash types, can lead to a large number of unique categories, resulting in increased computational complexity. Moreover, dealing with missing values in these structured features complicates the prediction task, which requires specialized handling to ensure robust model performance.

To tackle these challenges, we employ the target encoding method, also known as mean encoding or likelihood encoding \citep{micci2001preprocessing}. This technique enables us to convert categorical data into continuous numerical representations for seamless integration into the predictive model. The target encoding process involves mapping each high-cardinality categorical item to the probability estimate of the target variable (in this case, the CMF value). For CMF prediction, given a specific categorical circumstantial feature $\mathbf{c}$ (e.g., area type) with $N_{TR}$ samples in the training set and its corresponding CMF values, the target encoding computes the scalar encoding value $d_i$ for each unique categorical item $c_i$  (e.g., urban) in $\mathbf{c}$. The target encoding is computed using Equation \ref{eq.target_encoding}.

\begin{equation}
d_i = \lambda(n_i) \frac{\sum_{c\in c_i} cmf_{c}}{n_i} + (1-\lambda(n_i)) \frac{\sum_{c=1}^{N_{TR}} cmf_{c}}{N_{TR}}
\label{eq.target_encoding}
\end{equation}

where $d_i$ is the scalar encoding value for categorical item $c_i$,  $cmf_c$ is the CMF value at the observation cell where $c_i$ appears in the training set, $\lambda(n_i) \in [0, 1]$ is the weight factor that monotonically increases with $n_i$, defined as $\lambda(n_i) = \frac{n_i}{n_i + l}$, and $l$ is an empirical control factor to adjust the impact of the weight factor $\lambda$, with an empirical control factor $l=100$ in this study \citep{micci2001preprocessing}.

The target encoding approach effectively blends the posterior probability of CMF values for categorical item $c_i$ with the prior probability, considering both local and global information. Notably, the target encoding method treats missing values as any other value, like $c = c_0$. This ensures compatibility without requiring additional handling. If a missing value $c_0$  specific relevance for the CMF value, its scalar encoding $d_0$ captures that information; otherwise, $d_0$ represents a "neutral" representation with no extra implications.

\textbf{Input feature integration}. After embedding or encoding the countermeasure descriptions and other contextual factors, 
Then the embedded numerical model input $\mathbf{v}\in \mathbb{R}^{(m+k)}$, is suited for use in the subsequent regression model. 

\begin{equation}
    \mathbf{v} =[cms\_encoder(pseudo\_sent(s)); te\_encoder(\mathbf{u})]
        \label{eq.si2vi}
\end{equation}

\subsection{Regressive modeling}

For the second stage,  we utilize a regression module to establish the mapping between the contextualized countermeasure scenario representation and the investigated CMF values. Given the complexity of the relationships between CMF values and countermeasure scenario representations, the the multi-layer perception neural network (MLP) is chosen or solving such pattern-mining problem \citep{wang2003artificial}. To streamline the explanation, the MLP regression model in the context of estimating $cmf_i$ for countermeasure scenario $s_i$ is described in Equation \ref{eq.ann}:

\begin{equation}
\hat{cmf_i} = \sigma(\Theta_L^T \sigma(\Theta_{L-1}^T (\cdots (\sigma(\Theta_1^T \mathbf{v})) \cdots )))
\label{eq.ann}
\end{equation}
where $\sigma(\cdot)$ is the activation function (e.g., sigmoid function),  and $L=2$ is the number of layers. The input $\mathbf{v}$ is the vector with embed countermeasure scenario information, while $\hat{cmf}$ is the estimation of corresponding CMF values.

\section{Experiments}
\subsection{Baselines} 
In our experiments, we benchmarked the proposed approach against two baseline methods to highlight its efficiency:

\begin{itemize}

    \item \textbf{Non-encoding.} This simple yet effective strategy treats countermeasures and other conditional elements as categorical factors.   
    We directly calculate the distances of the variables extracted from the CMF Clearinghouse for each countermeasure scenario.  CMF values for unknown scenarios are predicted by averaging CMF values from their 10-nearest neighbors. While this non-encoding approach provides straightforward predictions, it overlooks the underlying semantic information in countermeasure descriptions, potentially limiting its ability to capture nuanced relationships.

    \item \textbf{Non-fine-tuning.} To underscore the importance of our fine tuning strategy, we used the pretrained Sentence-BERT model \textit{ all-mpnet-base-v2} without further fine tuning. This non-fine-tuned model highlights the significance of incorporating domain-specific knowledge and refining the model through fine-tuning. Comparing our approach to this baseline demonstrates the improved CMF predictions achieved through fine-tuning, better aligned with the specific requirements of road safety scenarios.

\end{itemize}

\subsection{Model evaluation}
\begin{figure}[!ht]
    \centering
    \includegraphics[width=0.8\textwidth]{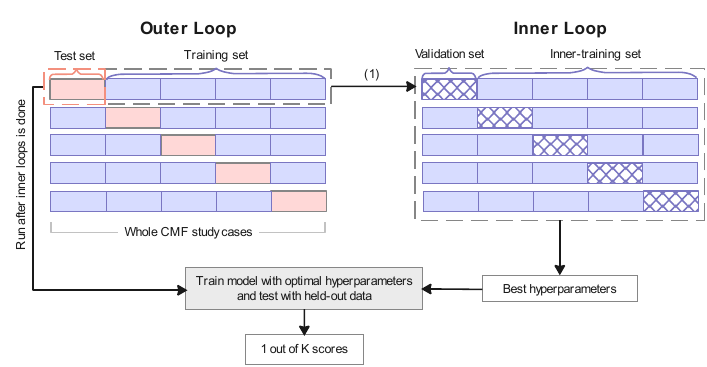}
    \caption{Nested cross-validation workflow for model evaluation}
    \label{fig:nestedcv}
\end{figure}

\textbf{Nested cross-validation.} To evaluate the performance of our predictive model, we adopt a nested cross-validation approach to ensure reliable evaluation, the workflow of which is demonstrated in Figure \ref{fig:nestedcv}. This involves an outer loop of 5-fold cross-validation, where the dataset is split into five equally-sized subsets: $D_1, D_2, D_3, D_4,$ and $D_5$. In each iteration of the outer loop, one subset is used as the test set, while the remaining four subsets are used for model fitting.
Within the inner loop of each iteration, we also perform a 5-fold cross-validation on the training data to select the optimal hyperparameters for our model. This is achieved using grid search with cross-validation. The best hyperparameters found during the inner cross-validation process are then used to train the model on the entire training data (excluding the test set from the outer loop). 
The trained model is then evaluated on the held-out test set for each iteration of the outer loop. This process is repeated five times, with each subset acting as the test set once. The final performance metrics are obtained by averaging the results from the five iterations, providing a robust assessment of the model's predictive capabilities.

\textbf{Evaluation metrics.} After training the model on each fold, we assess its performance by calculating the prediction errors on the corresponding test sets. The model performance metrics are then aggregated over the five folds to provide a comprehensive evaluation. 
To measure predictive accuracy, we employ two common metrics: the Mean Absolute Error (MAE) and the Root Mean Square Error (RMSE). These metrics offer valuable insights into the model's ability to align its predictions with the actual CMF values. 

Meanwhile, we introduce two additional evaluation metrics to further assess the model's performance given the road safety context. 
The Consistency Rate (CR) measures how many CMF predictions fall on the same side as the investigated CMFs (i.e., below or above 1.0). In practice, a CMF value less than $1.0$ indicates an expected reduction in crashes, while a CMF greater than $1.0$ represents an expected increase in crashes after the implementation of the target countermeasure. The CR metric provides important information about the model's ability to correctly predict whether a countermeasure leads to a reduction or an increase in crashes. Given that CMF values are reported with two decimal places to capture fine-grained crash outcome variations, it is essential to evaluate the model's precision to one decimal place. To achieve this, we introduce the Percentage of Precision (PoP) metric. PoP quantifies the model's ability to predict CMF values with the desired level of precision, considering the one-decimal place precision in CMF values.
The definitions of them are given by Equations \ref{eq:mae}-\ref{eq:pop}:

\begin{linenomath}
  \begin{flalign}
    MAE = \frac{1}{K}\sum_{j=1}^{K}\frac{1}{n_j}\sum_{i=1}^{n_j}|cmf_i^{(j)}-\hat{cmf_i}^{(j)}|
     \label{eq:mae}
  \end{flalign}
\end{linenomath}

\begin{linenomath}
  \begin{flalign}
    RMSE = \frac{1}{K}\sum_{j=1}^{K}\sqrt{\frac{1}{n_j}\sum_{i=1}^{n_j}(cmf_i^{(j)}-\hat{cmf_i}^{(j)})^2}
    \label{eq:rmse}
  \end{flalign}
\end{linenomath}

\begin{linenomath}
  \begin{flalign}
     CR = \frac{1}{K}\sum_{j=1}^{K}\frac{\sum_{i=1}^{n_j} \mathbb{I}\left((cmf_i^{(j)}-1)(\hat{cmf_i}^{(j)}-1) \geq 0\right)}{n_j}
    \label{eq:cr}
  \end{flalign}
\end{linenomath}

\begin{linenomath}
  \begin{flalign}
    PoP = \frac{1}{K} \sum_{j=1}^{K} \left( \frac{\sum_{i=1}^{n_j} \mathbb{I}\left(|cmf_{i}^{(j)} - \hat{cmf_{i}}^{(j)}| < 0.05\right)}{n_j} \times 100 \right)
    \label{eq:pop}
  \end{flalign}
\end{linenomath}

where $K=5$ is the number of folds, $n_j$ is the number of samples in the $j$-th fold, $cmf_i^{(j)}$ and $\hat{cmf_i}^{(j)}$ are the true and predicted CMF values in the $j$-th fold respectively.

\section{Results and discussion}
\subsection{Model performance in CMF prediction}
\subsubsection{Overall performance}

Table \ref{tab.performance} presents the comparison of the average model performance for CMF prediction across different testing folds in both roadway segments and intersections. The evaluation focuses on key metrics including MAE, RMSE, CR, and PoP. 
\begin{table}[!ht]
\centering
\caption{Summary of comparative model performance evaluation}
\label{tab.performance}
\resizebox{\textwidth}{!}{%
\begin{tblr}{
  row{1} = {c},
  cell{1}{1} = {r=2}{},
  cell{1}{2} = {c=4}{},
  cell{1}{6} = {c=4}{},
  cell{3}{6} = {c},
  cell{3}{7} = {c},
  cell{3}{8} = {c},
  cell{3}{9} = {c},
  cell{4}{6} = {c},
  cell{4}{7} = {c},
  cell{4}{8} = {c},
  cell{4}{9} = {c},
  cell{5}{6} = {c},
  cell{5}{7} = {c},
  cell{5}{8} = {c},
  cell{5}{9} = {c},
  hline{1,3,6} = {-}{},
  hline{2} = {2-9}{},
  hline{1,6} = {-}{0.08em},
}
Model        & Roadway       &               &               &               & Intersection  &               &               &               \\
             & MAE           & RMSE          & CR            & PoP           & MAE           & RMSE          & CR            & PoP           \\
Non-encoding & 0.18          & 0.25          & 0.81          & 43\%          & 0.22          & 0.30          & 0.78          & 37\%          \\
Non-tuning   & 0.16          & 0.23          & 0.84          & 46\%          & 0.17          & 0.26          & 0.82          & 45\%          \\
Our model    & \textbf{0.07} & \textbf{0.13} & \textbf{0.90} & \textbf{81\%} & \textbf{0.08} & \textbf{0.14} & \textbf{0.89} & \textbf{79\%} 
\end{tblr}
}
\end{table}

In our evaluation,we highlight the effectiveness of our proposed approach against baseline models.  For roadway scenarios,  the non-encoding model displayed an MAE of $0.18$ and an RMSE of $0.25$. After applying semantic encoding without fine-tuning (i.e. the non-tuning model), we  observed improvements with an MAE of $0.16$ and an RMSE of $0.23$. Notably, our proposed model, which incorporates both semantic encoding and fine-tuning, achieved the highest accuracy, with an MAE of $0.07$ and an RMSE of $0.13$.  The model also demonstrated highest consistency (CR) and precision (PoP), with 90\% consistency and 81\% precision in predictions. Similar performance trends were observed in intersection scenarios. The proposed model consistently outperformed baseline models, yielding an MAE of 0.08, an RMSE of 0.14, a CR of 0.89, and a PoP of 79\%. These results underscore the effectiveness of our approach in accurately estimating CMFs, particularly when considering the intricate contexts of roadway and intersection scenarios.

In addition, our model's performance remains consistent across roadway and intersection scenarios, despite differences in characters (sample size, independent variables, and missing values) of the corresponding data sets and underlying scenarios. This indicates the consistency of our model in making CMF predictions for various  infrastructure types and datasets, which is crucial when applied to new scenarios not covered by the FHWA clearinghouse.

\begin{figure}[!ht]
    \centering
    \includegraphics[width=1.0\textwidth]{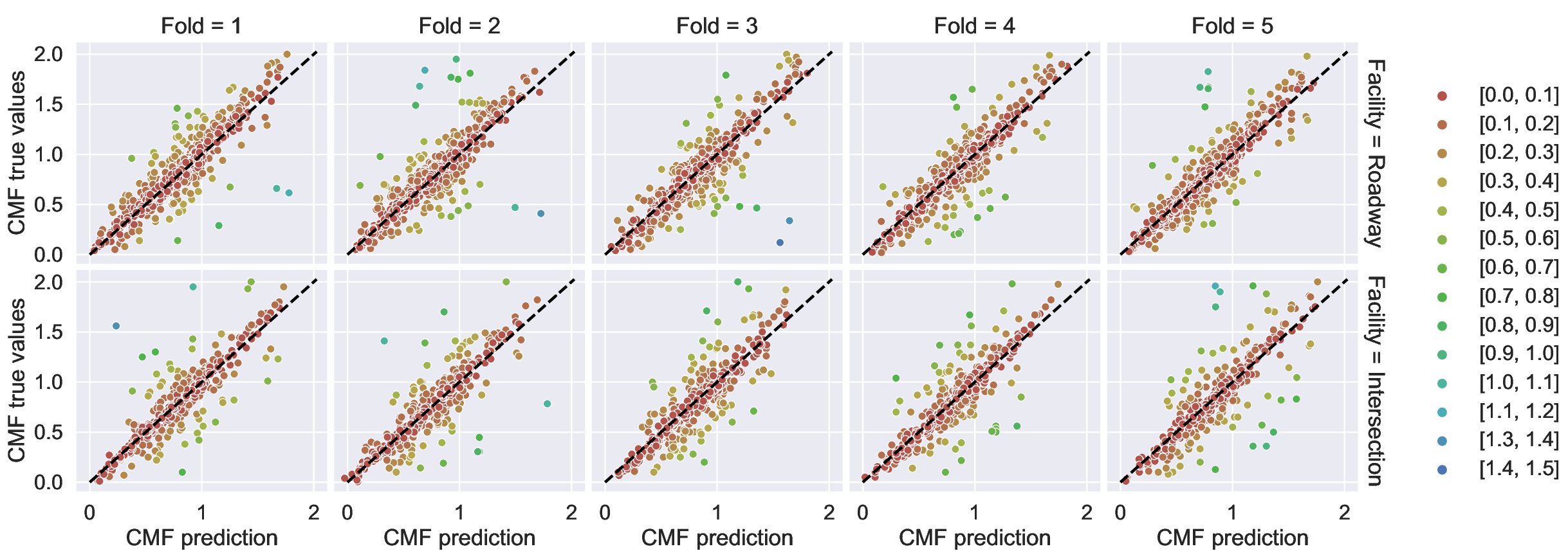}
    \caption{The scatter plots of the CMF predictions and the CMF true values on 5 folds for countermeasures under the roadway segment type (above) and the intersection type (below)}
    \label{fig:cv_scatter}
\end{figure}

Figure \ref{fig:cv_scatter} provides a visual comparison of CMF predictions against their ground-truth values for both intersections and roadway segments within each fold of the cross-validation process. Different colors are used to distinguish testing samples based on their absolute residual levels. The graphical representation demonstrates a clustering of data points symmetrically distributed around the diagonal line. This distribution indicates the overall accuracy and unbiased nature of the predictions, with only a minor presence of outliers observed across all folds.

\subsubsection{Predicting structured countermeasure scenarios}
\begin{figure}[!ht]
\begin{minipage}{0.48\textwidth}
\centering
\captionsetup{type=table}
\resizebox{\linewidth}{!}{%
    \begin{tabular}{ll}
    \hline
    \multicolumn{1}{c}{Structured shoulder-width treatment} \\ 
    \hline
    \fcolorbox{red}{white}{Widen} \fcolorbox{green}{white}{paved} shoulder from \fcolorbox{cyan}{white}{3} ft to \fcolorbox{orange}{white}{4} ft \\
    \fcolorbox{red}{white}{Widen} shoulder (\fcolorbox{green}{white}{paved}) (from \fcolorbox{cyan}{white}{0} to \fcolorbox{orange}{white}{4} ft) \\
    \fcolorbox{red}{white}{Widen} shoulder (\fcolorbox{green}{white}{unpaved}) (from \fcolorbox{cyan}{white}{0} to \fcolorbox{orange}{white}{2} ft) \\
    \fcolorbox{red}{white}{Pave} \fcolorbox{green}{white}{deteriorated} shoulder (\fcolorbox{cyan}{white}{2} ft) \\
    \fcolorbox{red}{white}{Reduce} \fcolorbox{green}{white}{paved} shoulder from \fcolorbox{cyan}{white}{3} ft to \fcolorbox{orange}{white}{1} ft \\
    \hline
    \end{tabular}
}
\caption{An example of structured countermeasures}
\label{tab:structure}
\end{minipage}%
\hfill
\begin{minipage}{0.48\textwidth}
\centering
\captionsetup{type=figure}
\resizebox{\linewidth}{!}{%
    \includegraphics{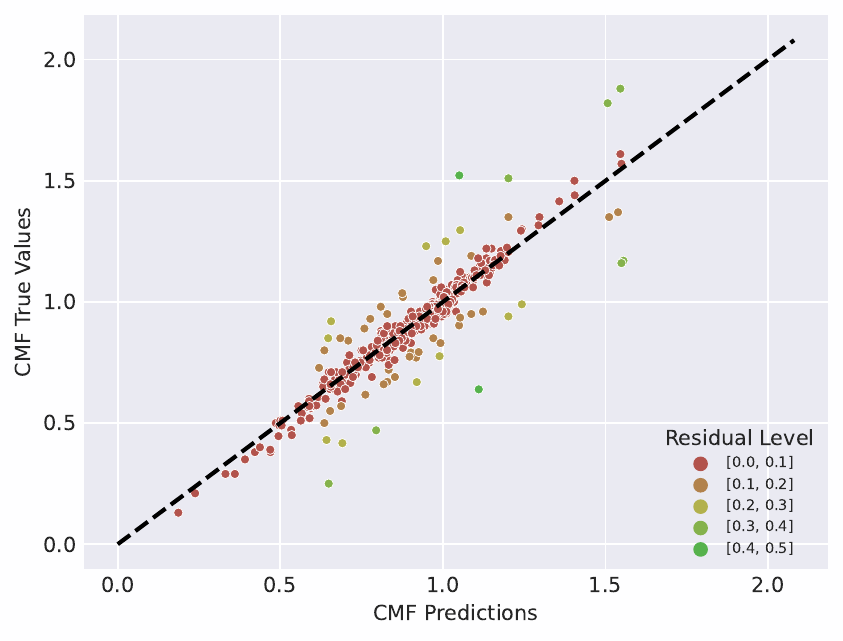}
}
\caption{CMF predictions on structured shoulder-width countermeasures}
\label{fig:structure_fig}
\end{minipage}
\end{figure}

As some countermeasures may involve quantitative alterations to existing facilities, these adjustments are often reflected in their descriptions. These descriptions consistently contain structured information related to a specific countermeasure category, such as the structured information contained within countermeasure descriptions related to shoulder width in Table \ref{tab:structure}, forming what we term \textit{structured countermeasures}.  

Our semantic encoding approach is fundamentally designed for general-purpose predictions, as approximately 90\% of countermeasures feature unstructured descriptions that lack explicit quantitative indicators. However, it is important to note that \textit{structured countermeasures} represent a specialized subset, and the challenge lies in the fact that these scenarios often exhibit minimal changes in their descriptions. If countermeasure scenario representations do not adequately capture the nuanced variations among them, CMF predictions may yield very similar values that deviate from the actual investigated CMF values.

To assess the performance of our model on \textit{ structured countermeasures}, we conducted a specific case study focusing on countermeasures associated with shoulder width adjustments. In this evaluation, we collect the prediction results of 330 structured countermeasures related to shoulder width adjustments from the CMF Clearinghouse repository.

The results, as visualized in Figure \ref{fig:structure_fig}, demonstrate a good alignment between the CMF predictions generated by our model and the corresponding investigated CMF values. This highlights the proficiency of our model in capturing the nuanced aspects of various countermeasure scenarios, enabling it to provide accurate CMF predictions. This is noteworthy given that our model is designed for mianly for unstructured data, but it achieves similar accuracy and precision even with structured data, demonstrating its robustness and adaptability.

\subsubsection{Predicting individual countermeasures and their combinations}
Our knowledge-mining-based CMF prediction model possesses a distinctive versatility that sets it apart. While it can provide initial estimations of the effects of individual countermeasures, it can also assess the combined impacts of these measures. This feature aligns closely with real-world road safety scenarios, where transportation authorities often deploy multiple safety interventions in tandem. The ability to predict the cumulative effects of these measures reflects the practical relevance of our approach.

\begin{figure}[!ht]
    \centering
	\includegraphics[width=0.6\textwidth]{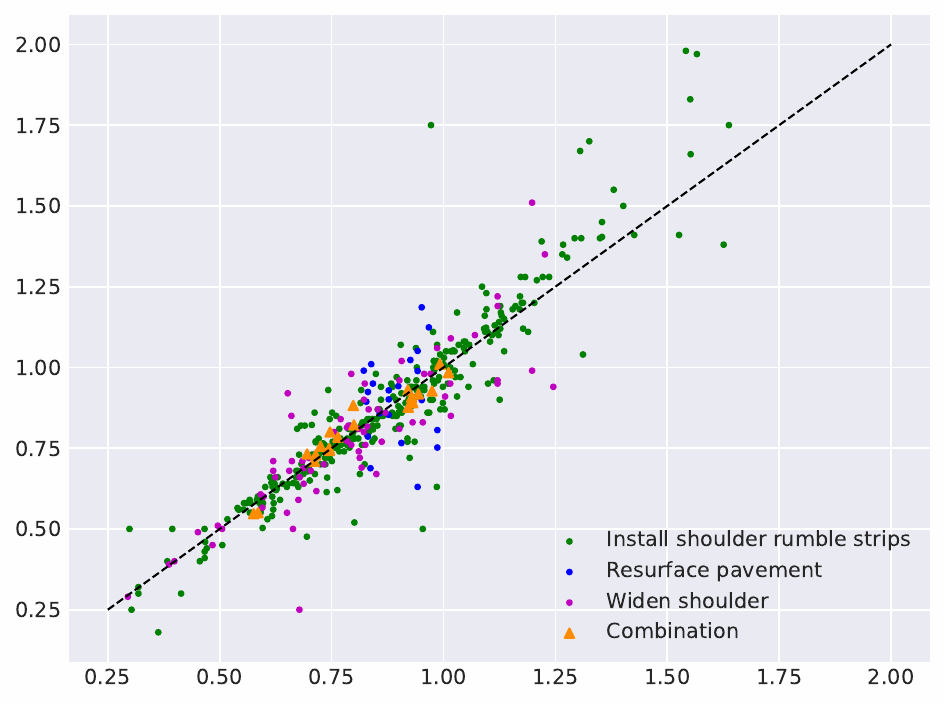}
    \caption{Model testing CMF predictions for individual countermeasures (\textit{install rumble strips}, \textit{resurface pavement}, and \textit{widen shoulder/widen shoulder (from 0 to 2 feet)}) and their combination (\textit{Install shoulder rumble stripe, widen shoulder from 0 to 2 feet, and pavement resurfacing})}
    \label{fig.combine}
\end{figure}

To illustrate the real-world applicability, we consider a scenari where a transportation authority is implementing a combination of countermeasures: \textit{install shoulder rumble stripes}, \textit{shoulder widening}, and \textit{resurface pavement}.  Our model is capable of predicting the individual impact of each countermeasure under different contextual situations. Furthermore, it can estimate the CMF when these measures are applied together. Notably, these three individual countermeasures and their combinations have been previously investigated and are included in the CMF Clearinghouse. Figure \ref{fig.combine} demonstrates our model's ability to provide initial estimations for these different individual countermeasures (represented as dot markers) and the combination of these measures into an integrated countermeasure strategy (represented as triangular markers). This dual prediction capability informs authorities of the combined effect of multiple individual measures, facilitating informed initial decisions. 

\subsection{Analysis of CMF prediction results}
To identify how these outliers come from, we first conducted a detailed analysis to identify countermeasure categories with relatively accurate predictions and those requiring further investigation. Evaluation metrics are presented in Table~\ref{tab:roadway_cat} (roadway segments) and Table~\ref{tab:intersec_cat} (intersections). 

\begin{table}[!ht]
\centering
\caption{CMF testing prediction evaluation results for roadway segment}
\label{tab:roadway_cat}
\begin{tblr}{
  width = \linewidth,
  colspec = {Q[398]Q[75]Q[96]Q[69]Q[69]Q[227]},
  hline{1,19} = {-}{0.08em},
  hline{2} = {-}{},
}
Countermeasure Category     & MAE  & RMSE & CR   & PoP  & Instance Number \\
Highway lighting            & 0.04 & 0.07 & 1.00 & 92\% & 60              \\
Railroad grade crossings    & 0.05 & 0.08 & 1.00 & 91\% & 11              \\
Delineation                 & 0.05 & 0.10 & 0.85 & 92\% & 275             \\
Advanced technology and ITS & 0.06 & 0.10 & 0.98 & 87\% & 272             \\
Shoulder treatments         & 0.06 & 0.11 & 0.89 & 92\% & 1306            \\
Signs                       & 0.06 & 0.13 & 0.98 & 89\% & 132             \\
Speed management            & 0.07 & 0.11 & 0.82 & 87\% & 186             \\
Alignment                   & 0.07 & 0.11 & 0.94 & 82\% & 85              \\
Roadway                     & 0.07 & 0.13 & 0.92 & 91\% & 1474            \\
Roadside                    & 0.07 & 0.14 & 0.88 & 86\% & 598             \\
Work zone                   & 0.08 & 0.12 & 0.94 & 81\% & 80              \\
Interchange design          & 0.08 & 0.14 & 0.95 & 80\% & 60              \\
Access management           & 0.08 & 0.15 & 0.96 & 80\% & 349             \\
Pedestrians                 & 0.09 & 0.22 & 0.95 & 85\% & 41              \\
On-street parking           & 0.10 & 0.21 & 0.96 & 82\% & 28              \\
Bicyclists                  & 0.11 & 0.16 & 0.88 & 63\% & 110             \\
Transit                     & 0.19 & 0.31 & 0.92 & 58\% & 12              
\end{tblr}
\end{table}

\begin{table}[!ht]
\centering
\caption{CMF testing prediction evaluation results for intersection}
\label{tab:intersec_cat}
\begin{tblr}{
  width = \linewidth,
  colspec = {Q[390]Q[73]Q[94]Q[67]Q[85]Q[223]},
  hline{1,18} = {-}{0.08em},
  hline{2} = {-}{},
}
Countermeasure Category    & MAE  & RMSE & CR   & PoP   & Instance Number   \\
Delineation                  & 0.04 & 0.06 & 0.98 & 96\%  & 45              \\
On-street parking            & 0.04 & 0.04 & 1.00 & 100\% & 2               \\
Highway lighting             & 0.05 & 0.07 & 0.86 & 93\%  & 44              \\
Signs                        & 0.05 & 0.09 & 0.91 & 85\%  & 80              \\
Roadway                      & 0.07 & 0.12 & 0.95 & 75\%  & 88              \\
Pedestrians                  & 0.08 & 0.12 & 0.97 & 87\%  & 70              \\
Intersection traffic control & 0.08 & 0.14 & 0.89 & 94\%  & 1161            \\
Railroad grade crossings     & 0.09 & 0.13 & 1.00 & 67\%  & 3               \\
Advanced technology and ITS  & 0.09 & 0.14 & 0.91 & 78\%  & 196             \\
Intersection geometry        & 0.09 & 0.17 & 0.73 & 94\%  & 915             \\
Speed management             & 0.10 & 0.14 & 0.82 & 70\%  & 99              \\
Interchange design           & 0.10 & 0.17 & 0.90 & 76\%  & 78              \\
Access management            & 0.10 & 0.20 & 0.88 & 72\%  & 92              \\
Transit                      & 0.12 & 0.16 & 0.96 & 52\%  & 27              \\
Bicyclists                   & 0.19 & 0.27 & 0.67 & 54\%  & 63              \\
Roadside                     & 0.19 & 0.32 & 0.71 & 71\%  & 7               
\end{tblr}
\end{table}

Notably, Certain countermeasure categories, such as ``Highway lighting," ``Railroad grade crossings," and ``Delineation"  consistently exhibit excellent prediction results in both roadway segments and intersections, with high CR and PoP rates, demonstrating accurate and consistent predictions. On the other hand, both the road and intersection countermeasures in the categories of `Bicyclists" and ``Transit" consistently show relatively higher MAE values ($0.11, 0.19$ and $0.19, 0.19$, respectively). These two categories also show lower PoP values, indicating a moderate precision in predicting CMF values. The PoP values for ``Bicyclists" (63\% and 54\%) and ``Transit" (58\% and 52\%) are significantly lower than other categories, and their RMSE values are also relatively larger. 
This indicates that these two countermeasure categories contribute the most to the outliers in Figure~\ref{fig:cv_scatter}. For transit and pedestrian-related crashes, they can involve more complex interactions between passengers, cyclists, and motorists, which may make it more challenging to generate accurate predictive results on the effectiveness of the corresponding countermeasures. There are also more obvious data sparsity problems for the ``Transit" and `Roadside" categories, as countermeasures are less studied and have limited data available in comparison to more common countermeasures like roadway configurations or traffic control measures, with only a few countermeasures investigated under such extensive general category extents.

\begin{figure}[!ht]
    \centering
    \subfigure[Subgroup MAE distribution for roadway-related countermeasures]{
        \includegraphics[width=3.8in]{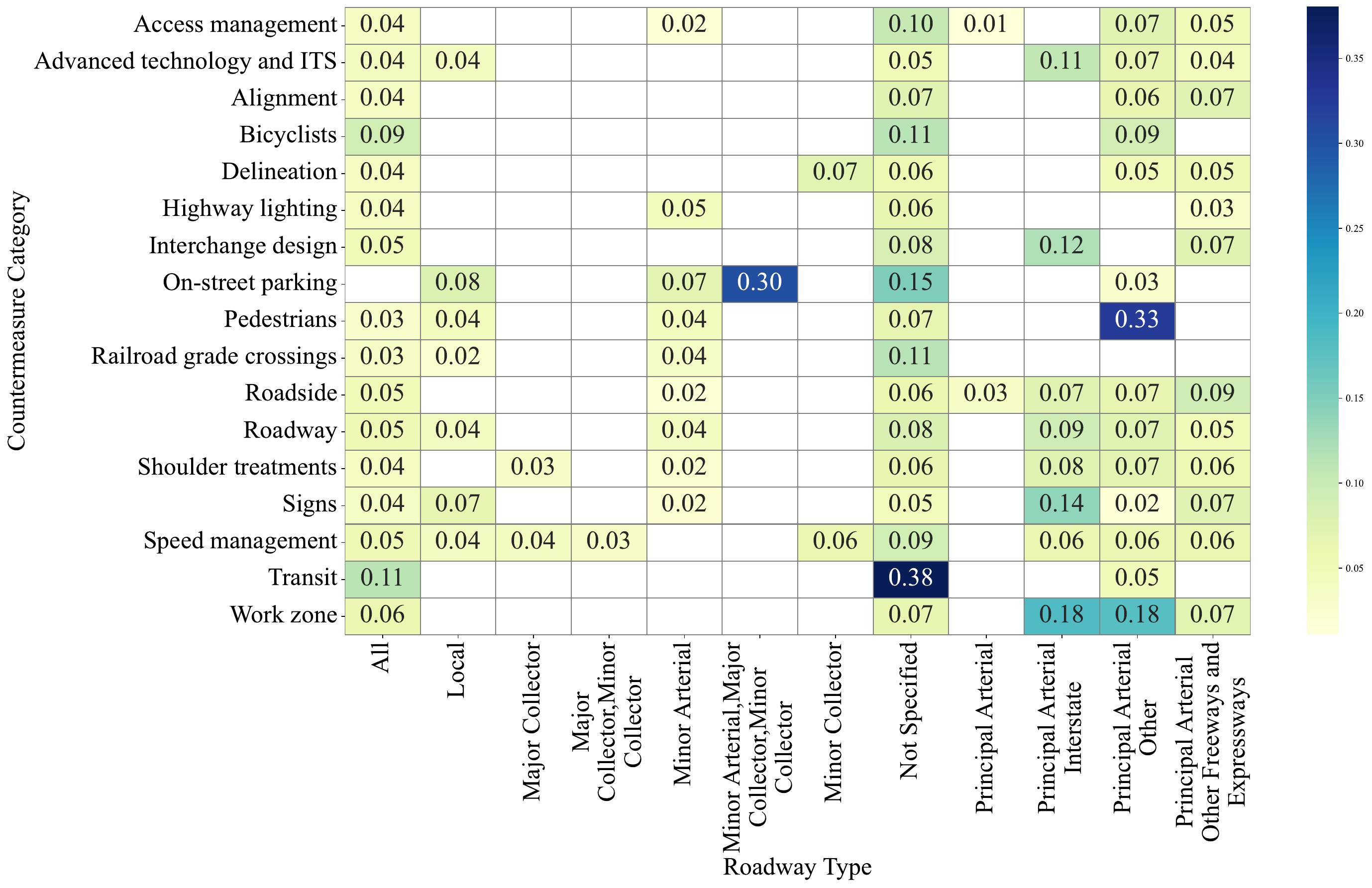}
    }
    \subfigure[Subgroup MAE distribution for intersection-related countermeasures]{
	\includegraphics[width=3.8in]{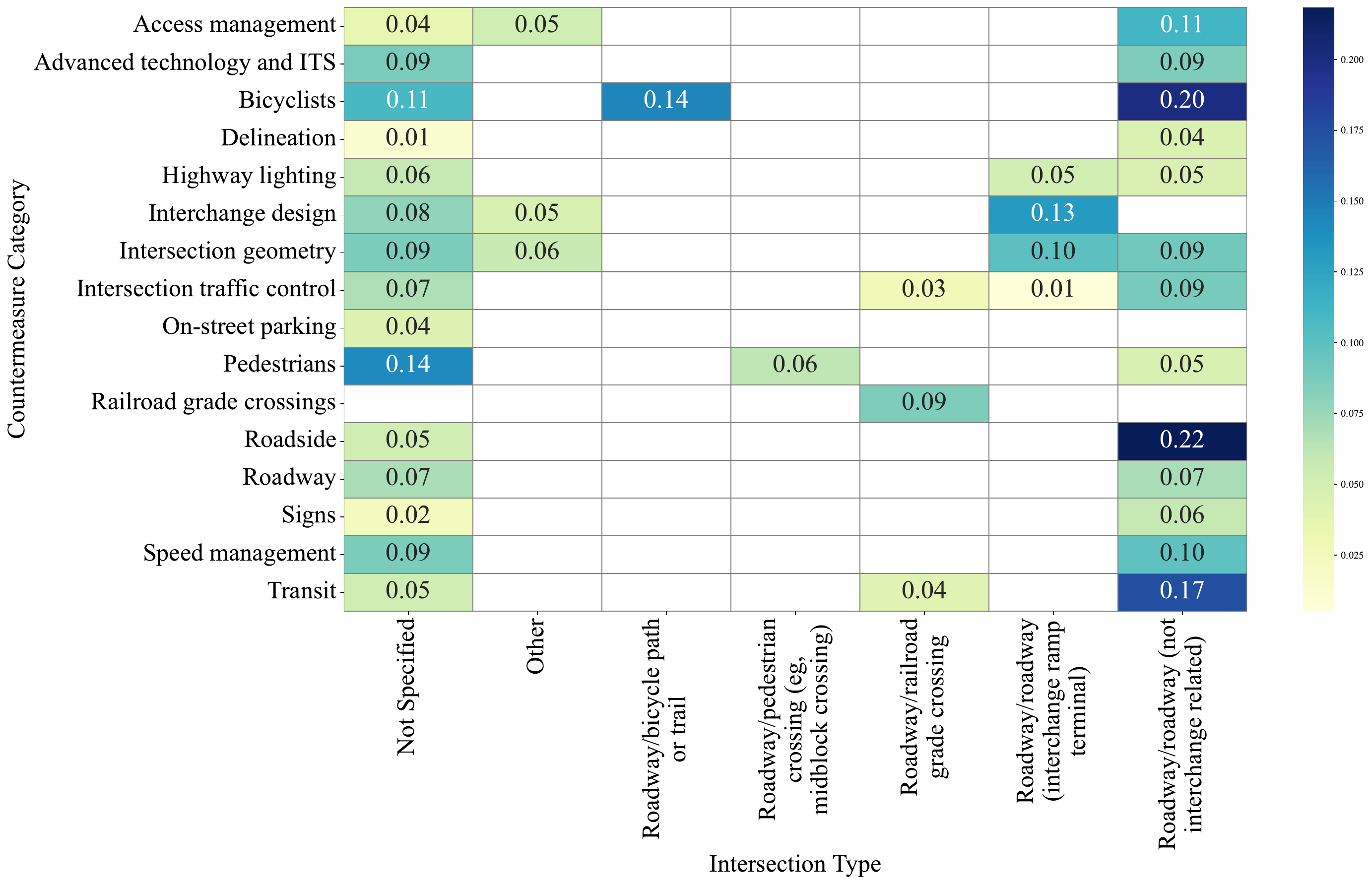}
    }
    \caption{MAE distribution for subgroups based on site-condition types and countermeasure categories: roadway segment and intersection type. Lighter colors indicate smaller MAEs.}
    \label{fig.heatmap}
\end{figure}

We further conducted an analysis by grouping test samples into various subgroups based on infrastructure types and countermeasure categories. The results are shown in Figure \ref{fig.heatmap}, presenting the MAE values for each subgroup.

Upon examining the distribution of subgroup MAEs, we observed that the majority of subgroups exhibited MAE values closely aligned with the overall test MAE for both intersection and roadway segment types. This indicates that our model has a consistent performance across different scenarios. However, certain cells contain higher MAE values, which may shed light on potential sources of outliers  observed in Figure~\ref{fig:cv_scatter}. In roadway-related scenarios, anomalies were noted in the MAE values of specific subgroups, including ``Transit" countermeasures with an unspecified roadway type, ``On-street parking" countermeasures on the combined site-conditions of  ``Minor Collector" roadway type, and ``Pedestrian" countermeasures on the ``Principal Arterial Other" roadway type, with MAE values of $0.38$, $0.30$, and $0.33$, respectively. Similarly, in intersection-related scenarios, anomalies were observed in subgroups of ``Roadside", ``Bicyclists", and ``Transit" countermeasures on the ``Roadway/Roadway (not interchange related)" intersection type, with MAE values at $0.22$, $0.20$, and $0.17$, respectively.  In particular, the lower prediction accuracy for the ``Transit'' and ``Bicyclist'' categories is primarily associated with samples from the intersection scenario.

\subsection{Model limitations}

While our knowledge-mining approach has shown promise and has the potential to enhance the estimation of CMF in the field of road safety, it is crucial to acknowledge the inherent limitations of our methodology. One key aspect to consider is the data coverage within the CMF Clearinghouse, which forms the backbone of our model's knowledge. Although the clearinghouse provides a wealth of valuable data, it may not comprehensively represent all potential countermeasure scenarios' variation patterns or or reflect the full breadth of contextual factors. Although the clearinghouse provides a wealth of valuable data, it may not capture the full spectrum of variation patterns among potential countermeasure scenarios or encompass the entire breadth of contextual factors. Simultaneously, inaccuracies, missing values, and inconsistencies that are inherent in the existing, case-specific CMF studies can introduce noise into our model, affecting its overall performance.  This limitation can impact the model's accuracy in scenarios that significantly deviate from the available data. However, we view this as an opportunity for future collaborations and data expansion efforts. By closely partnering with transportation authorities and safety researchers, we can enrich the clearinghouse with a broader array of countermeasure scenarios, thereby enhancing our model's capacity to address a wider range of safety interventions.

Another noteworthy limitation pertains to the adaptability of our model to temporal changes in road safety conditions. As our predictions rely on historical data, they may not fully account for the evolving nature of countermeasure effectiveness over time. This challenge highlights the importance of ongoing model updates and evaluations, enabling us to incorporate temporal considerations and adjust our predictions as needed. Furthermore, while our model exhibits promise in predicting CMFs for various countermeasures, we recognize the potential need for additional testing and validation through the lens of human expert knowledge. As with any innovative approach, the generalization capabilities of the model require continuous refinement to ensure its effectiveness in meeting the diverse needs of road safety professionals and decision-makers. Viewing these limitations through a constructive lens reveals opportunities for future refinement and expansion of our knowledge-mining approach, which ultimately contributes to safer roadways and more informed decision-making in the realm of transportation safety.

\section{Conclusions}

In the domain of road safety, the estimation of CMFs plays a pivotal role in assessing the effectiveness of diverse safety countermeasures under varying conditions. Traditional methods, however, often fall short when faced with the ever-evolving landscape of these countermeasure scenarios, highlighting the need for a more adaptable solution. The conventional approaches to CMF estimation primarily rely on crash data and are tailored to specific cases, rendering them expensive and less flexible in scenarios with limited data availability.

Our approach draws on the extensive knowledge on CMF contained within the FHWA CMF Clearinghouse. It enables the prediction of CMFs for countermeasure scenarios, even when detailed crash data is scarce or time is of the essence. 
Another central contribution of our research lies in the data-driven nature of our knowledge-mining approach. By effectively capturing the semantic context associated with diverse countermeasure scenarios, our framework guarantees compatibility with varying data types, including cases with missing information. This adaptability is critical due to the wide spectrum of safety countermeasure scenarios, each of which exhibits its unique characteristics and possible missing contextual conditions.

Our experimental evaluation with CMF Clearinghouse data validates the effectiveness of our approach in predicting CMFs with reasonable accuracy. The results demonstrate that our approach offers a practical and efficient solution that enables safety professionals to make informed decisions, even in scenarios characterized by limited data availability and a wide array of countermeasure scenarios. As we tap into the wealth of existing CMF knowledge, our framework contributes to enhancing road safety and supporting decision-making in the ever-changing and dynamic environment.

Moreover, we anticipate several potential extensions to our work. First, combining our approach with traditional methods may promise enhanced CMF estimation accuracy while lowering costs. Second, incorporating domain-specific knowledge could further enhance the model's predictive capabilities, such as when dealing with countermeasures sharing similar effects but falling into different categories.  Lastly, as the CMF Clearinghouse continues to accumulate more case-specific CMF records, our model stands to benefit from the additional information, leading to improved performance.

\section{Declaration of competing interest}
The authors declare that they have no known competing financial interests or personal relationships that could have appeared to influence the work reported in this paper.

\section{Acknowledgements}
This research is supported by California Senate Bill-1 and Caltrans grants ``Developing a Safety Effectiveness Evaluation Tool for California (Phase I and II)'', and Center for Transportation, Environment, and Community Health, a USDOT Tier 1 University Center.

\section{Data availability}
The data are available at \url{https://www.cmfclearinghouse.org/results.php}.

\bibliography{mybibfile}

\end{document}